\PassOptionsToPackage{unicode}{hyperref}
\PassOptionsToPackage{hyphens}{url}
\documentclass[11pt]{article}
\usepackage[textwidth=6in,top=1in,bottom=1in]{geometry}
\usepackage{float}
\usepackage{microtype}
\usepackage{xurl}
\usepackage[font={small,it},labelfont=bf,skip=8pt]{caption}
\usepackage{xcolor}
\usepackage{amsmath,amssymb}
\usepackage{iftex}
\ifPDFTeX
  \usepackage[T1]{fontenc}
  \usepackage[utf8]{inputenc}
  \usepackage{textcomp} % provide euro and other symbols
\else % if luatex or xetex
  \usepackage{unicode-math} % this also loads fontspec
  \defaultfontfeatures{Scale=MatchLowercase}
  \defaultfontfeatures[\rmfamily]{Ligatures=TeX,Scale=1}
\fi
\usepackage{lmodern}
\usepackage{newunicodechar}
\newunicodechar{−}{\ensuremath{-}}
\newunicodechar{γ}{\ensuremath{\gamma}}
\newunicodechar{∆}{\ensuremath{\Delta}}
\newunicodechar{Δ}{\ensuremath{\Delta}}
\newunicodechar{→}{\ensuremath{\rightarrow}}
\newunicodechar{×}{\ensuremath{\times}}
\newunicodechar{≈}{\ensuremath{\approx}}
\newunicodechar{≥}{\ensuremath{\geq}}
\newunicodechar{≤}{\ensuremath{\leq}}
\newunicodechar{Ł}{\L}
\ifPDFTeX\else
\fi
\IfFileExists{upquote.sty}{\usepackage{upquote}}{}
\IfFileExists{microtype.sty}{% use microtype if available
  \usepackage[]{microtype}
  \UseMicrotypeSet[protrusion]{basicmath} % disable protrusion for tt fonts
}{}
\makeatletter
\@ifundefined{KOMAClassName}{% if non-KOMA class
  \IfFileExists{parskip.sty}{%
    \usepackage{parskip}
  }{% else
    \setlength{\parindent}{0pt}
    \setlength{\parskip}{6pt plus 2pt minus 1pt}}
}{% if KOMA class
  \KOMAoptions{parskip=half}}
\makeatother
\usepackage{longtable,booktabs,array}
\usepackage{caption}
\usepackage{calc} % for calculating minipage widths
\usepackage{etoolbox}
\makeatletter
\patchcmd\longtable{\par}{\if@noskipsec\mbox{}\fi\par}{}{}
\makeatother
\IfFileExists{footnotehyper.sty}{\usepackage{footnotehyper}}{\usepackage{footnote}}
\makesavenoteenv{longtable}
\usepackage{graphicx}
\makeatletter
\newsavebox\pandoc@box
\newcommand*\pandocbounded[1]{% scales image to fit in text height/width
  \sbox\pandoc@box{#1}%
  \Gscale@div\@tempa{\textheight}{\dimexpr\ht\pandoc@box+\dp\pandoc@box\relax}%
  \Gscale@div\@tempb{\linewidth}{\wd\pandoc@box}%
  \ifdim\@tempb\p@<\@tempa\p@\let\@tempa\@tempb\fi% select the smaller of both
  \ifdim\@tempa\p@<\p@\scalebox{\@tempa}{\usebox\pandoc@box}%
  \else\usebox{\pandoc@box}%
  \fi%
}
\def\fps@figure{H}
\makeatother
\providecommand{\tightlist}{%
  \setlength{\itemsep}{4pt}}
\usepackage{bookmark}
\IfFileExists{xurl.sty}{\usepackage{xurl}}{} % add URL line breaks if available
\makeatletter
\@ifundefined{xmpquote}{}{}
\makeatother
\hypersetup{
  pdftitle={Increasing Skill Level Recruits Deeper Attention Layers in a Frozen Chess Transformer},
  pdfauthor={David Litman},
  hidelinks,
  pdfcreator={LaTeX via pandoc}}

\title{Increasing Skill Level Recruits Deeper Attention\\ Layers in a Frozen Chess Transformer}
\author{David Litman\\[2pt] \small Computational Neurobiology Laboratory, Salk Institute}
\date{}

\begin{document}
\setlength{\LTpre}{16pt}\setlength{\LTpost}{16pt}\setlength{\intextsep}{12pt}\setlength{\parskip}{9pt plus 2pt minus 1pt}\Urlmuskip=0mu plus 1mu\relax
\maketitle

\begin{center}\textbf{Abstract}\end{center}\vspace{-6pt}
\begin{list}{}{\setlength{\leftmargin}{1cm}\setlength{\rightmargin}{1cm}}\item[]
\addcontentsline{toc}{section}{Abstract}

Chess involves complex reasoning in a deterministic environment, which makes it a useful setting for studying the mechanisms of computation inside transformers. The Maia-3 chess transformer takes Elo,
a measure of competitive chess skill, as an input to the pre-trained network, so we can vary the skill the network is conditioned on with no change to its weights. Here we investigate how turning this
skill dial affects self-attention. Ablating every attention head at every Elo from 700 to 2500, we find 1) increasing skill pushes the causal center of mass of the computation deeper, monotonically,
for every chess piece and move type we measured; 2) the depth migration is much greater for specific tactics, especially knight forks, than for other move types; 3) the migration consists of deeper
heads getting recruited for more specialized computations while one shared shallow head keeps a roughly constant contribution. These results may shed light on how conditioning inputs redistribute
computation in larger transformers.

\end{list}
\vspace{6pt}
\section{Introduction}\label{introduction}

Chess has long been one of the most popular areas of AI research and innovation (Silver et al., 2017), in large part because the game necessitates the sorts of judgment and complex reasoning that the
field of AI seeks to imbue computers with (Shannon, 1950). It also carries a hierarchy of human concepts, from squares to threats to tactics to strategic considerations, to compare against a model's
internal feature space. The domain comes equipped with centuries of diverse human data about chess theory and decision making to train on as well. The past few decades of research have led to the
development of superhuman models such as DeepMind's tabula rasa AlphaZero program (Silver et al., 2017). Other well-known models include a large transformer (Vaswani et al., 2017) trained to
grandmaster-level skill from its policy output alone, with no search (Ruoss et al., 2024), and the industry-standard engine Stockfish, whose search is guided by a small neural network that evaluates
positions using a sparsely activated set of tens of thousands of human-picked features. The Maia-3 model that we report on belongs to a newer line of models trained to mimic human rather than
superhuman play, with all of our tendencies and biases (McIlroy-Young et al., 2020; Tang et al., 2024; Monroe et al., 2026).

Our work adds to prior results on where concepts are represented in a network (Section \ref{related-work}), but differs in that it holds every weight and bias fixed and adjusts a single input
variable, whereas those experiments compare different models or different training checkpoints of the same model. We are not aware of prior interpretability work that does this to ask where
computation happens. One immediate takeaway is that even in relatively small models, circuits may not stay where they were first observed under one condition. Another is that the direction of the
migration in our chess network echoes a finding about human players, whose planning depth grows with expertise (van Opheusden et al., 2023). How far our results generalize to other settings, such as
raising the reasoning effort of a large language model, is a topic for future work.

\begin{figure}
\centering
\makebox[\linewidth][c]{\includegraphics[width=7.5in,keepaspectratio]{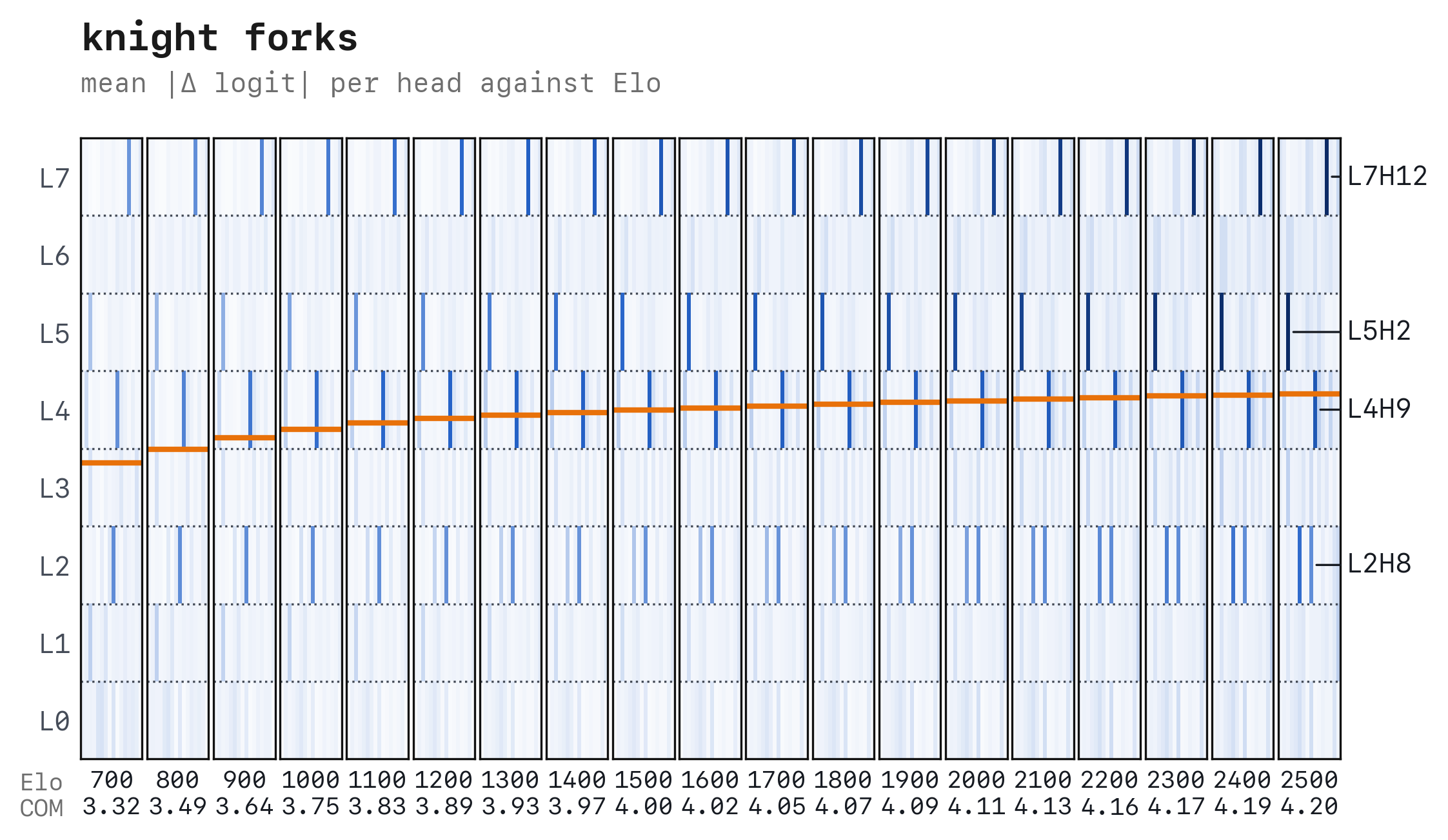}}
\caption{The causal center of mass for knight forks deepens with increasing Elo. Figure described in detail in Section \ref{results}.}\label{fig:ladder-knight}
\end{figure}

\section{Background}\label{background}

\subsection{Mechanistic interpretability}\label{mechanistic-interpretability}

As the field is relatively new, it is worth taking some time to give extra context to unfamiliar readers. The aim of mechanistic interpretability is to better understand how AI works by
reverse-engineering models into human-describable parts (Olah et al., 2020). The major hope is that this will help safety and alignment by allowing us to recognize misaligned model internals before it
produces misaligned behavior. In this field, chess offers what language cannot: there is an enumerable legal-move set and every position can be run in strong chess engines to provide a ground truth
for model outputs.

\subsection{Terminology we will keep constant throughout}\label{terminology-we-will-keep-constant-throughout}

\begin{itemize}
\tightlist
\item
  \textbf{layer 0--7} --- the eight transformer blocks; each contains an attention sublayer and an MLP sublayer.
\item
  \textbf{LmHn} --- the convention denoting: layer m, head n.
\item
  \textbf{residual stream} --- the shared communication channel that attention and MLP sublayers additively read from and write into.
\item
  \textbf{readout point} --- the residual stream's content at a given depth. We read at 18 points: the input embedding, the post-attention and post-MLP points of each of the 8 blocks, and the final
  encoder output.
\item
  \textbf{logit} --- the raw score the network gives each move before scores are turned into probabilities. A non-linear function over all 4352 possible moves gives the policy, so a logit for a move
  has meaning independent of the status of alternative moves. Ablation effects in this paper are measured by the change in a single move's logit.
\item
  \textbf{logit lens} --- a mechanistic interpretability technique to decode the residual stream at an intermediate layer by pushing it through the unembedding map as if it were the finished product
  to obtain logits.
\item
  \textbf{policy} --- the network's output distribution over all 4352 moves, also used for the output of a single move. Maia plays directly from this; there is no search.
\item
  \textbf{ablation} --- subtract an attention head's output on a given position from the residual stream, and measure how the model's output changes.
\item
  \textbf{check} --- a chess move that attacks the enemy king. A check has to be addressed immediately, which is why a fork wins material: the king has to address the check and the other attacked
  piece is simply captured the next move.
\item
  \textbf{attacked, defended, hanging} --- a piece is attacked if an enemy piece could capture it next move, defended if you could recapture on that square, and hanging if it is attacked but not
  defended, meaning it is free to take.
\item
  \textbf{knight fork} --- the knight moves in a unique ``L'' shape and can jump over any piece, which makes it the piece that most easily carries out forks: a single move that attacks two valuable
  pieces at once. In this paper we refer only to the most potent forks, those that attack the king and the queen and so win the queen.
\item
  \textbf{causal center of mass} --- (COM) a metric for the average depth at which computation happens. Given a move, measure the absolute change in its logit when ablating every head in the model
  individually to get their causal masses. Then compute the mass-weighted average layer (the sum of the layer number times mass in that layer divided by the total mass).
\end{itemize}

\subsection{The model}\label{the-model}

\textbf{Maia-3} is a transformer based chess model (chessformer) built to mimic human play across skill levels.

It is a bidirectional encoder over the 64 board squares. The input board enters as 64 square-tokens each representing the content of the square, and a single forward pass returns the full policy plus
a win/draw/loss estimation. Each token maps to one unique square, so attention maps are literally visualizable on the board. Its attention heads combine two schemes: the standard ``semantic''
self-attention, and a ``geometric'' GAB attention scheme which carries positional structure because chess pieces move in ways that complicate semantic only attention (Monroe et al., 2026).

\textbf{Skill conditioning.} ``Elo'' is a scalar that is used across the world to estimate a human player's competitive chess strength. In the US Chess Federation, an intermediate player might have an
Elo of 1200 and a local champion might have an Elo of 2100, though online ratings and data like ours skew ratings significantly higher. In the model, the variable Elo rating \emph{k} is a linear
combination of two learned 128-d vectors: \[e_k = γ*e_{weak} + (1 − γ)*e_{strong}, γ = \frac{(5000−k)}{5000}\] This is combined with the 64 square tokens for the forward pass. What is especially
notable is that the skill conditioning is a walk along a single straight line in the high dimensional space of the model. One can move along it while keeping every learned weight and bias frozen and
force the network to reorganize.

\subsection{Related work}\label{related-work}

\textbf{Chess Interpretability}

McGrath et al.~(2022) published one of the first ``chess interpretability'' papers: ``Acquisition of Chess Knowledge in AlphaZero.'' They sought to compare new mechanistic interpretability findings in
language models to the aforementioned AlphaZero which served as an ideal independent test of those findings since it was never trained on human data, controlling for the most complicated confounds for
why AI models tended to contain human interpretable features. They elegantly used simple probes that read a concept out of a layer's activations across both depth of the model and across checkpoints
of training to produce ``What When Where'' plots. These plots showed, for example, that the concept of the ``bishop pair'' is represented in the network the most clearly in very late training points
in middle to late layers. In other work, Adam Karvonen (2024) showed that the board itself is represented and used for computations even on the ``Chess-GPT'' model which was trained only on the
strings of the ordered moves in a game. Jenner et al.~(2024) found that Leela (another chessformer with a square-to-token scheme) learns a looking ahead computation and they identify piece-specific
attention heads. Litman (2026c) used similar ablation techniques to this paper and examined a single attention head in the 5 million parameter Maia-3 model and its distinctive role in suppressing
``aimless'' sorts of knight moves.

\textbf{Depth of computation}

Hu, Zhou and Zhang (2025) studied the Qwen-2.5 family of models and report that effective depth is less tied to capability than one might expect. Concretely, they show that when comparing a model to
its smaller distilled counterpart, there is little difference in where the computations occur. Also, that increasingly difficult reasoning problems on a math exam database do not activate deeper
layers for their tested reasoning model. These results are surprising but there is a difference in measuring how difficulty affects depth and how our variable of conditioning affects depth. In fact,
we do find harmonious evidence that tactical puzzle difficulty does not order the migration.

\subsection{Contributions}\label{contributions}

\textbf{We show that} turning Maia-3's Elo input from 700 to 2500 moves the causal center of mass of its attention heads deeper: +0.52 to +0.88 layers, monotonically, at 18/18 Elo steps for all four
forking piece types (p \textless{} 0.001, permutation test).

\textbf{We demonstrate} a methodology for studying a conditioning input on a frozen network with mechanistic interpretability tools.

\textbf{We release} chessformer\_lens: an open source library and interactive app for understanding chess transformers. It provides clean tooling for logit lens, ablation heatmaps, attention/GAB
decomposition, attention atlases and much more, with which every figure in this paper was generated (Litman, 2026d; DOI \href{https://doi.org/10.5281/zenodo.21877655}{10.5281/zenodo.21877655}).

\section{Methods}\label{methods}

\textbf{Head ablation sweeps.} Over a stack of input positions, we ablate each of the model's 8×16 attention heads in turn and record the change it makes to the logit of the move of interest.
Repeating this at every Elo input gives, per position, a tensor of shape (Elo, layer, head). Its mean over the positions is an 8×16 grid of logit changes at each Elo.

\textbf{Error bars.} Unless stated otherwise, every ± in this paper is a 95\% confidence interval.

\textbf{Mining the positions.} Every position comes from a Lichess.org database of human games played. We use predicates in Python to label these moves. A fork is a non-capturing move after which the
landing square is adequately defended and the piece attacks both a square occupied by the enemy king and a square occupied by the enemy queen. Forks in general need not involve the king and queen, but
we restrict to these ``royal'' forks. We only considered pawn, bishop, knight, and rook forks because a king is very fragile and can only fork in rare specific circumstances and a queen fork is not
usually a cleanly winning move because it results in a trade of equal material, queen for queen.

\begin{figure}
\centering
\makebox[\linewidth][c]{\includegraphics[width=\linewidth,keepaspectratio]{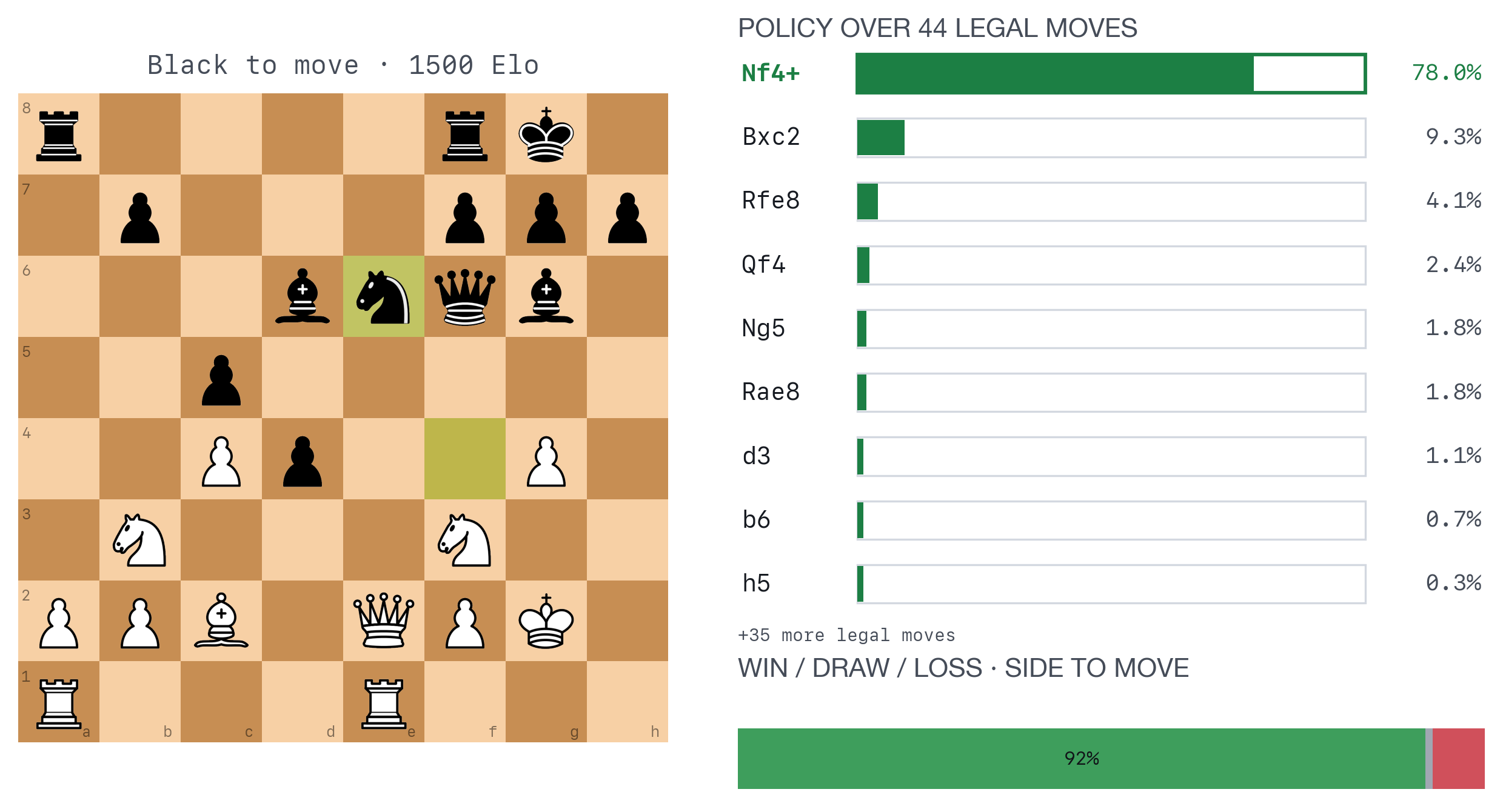}}
\caption{A running example of a knight fork and the Maia-3 output at 1500 Elo---the knight moves to square f4 to simultaneously attack the king and the queen.}\label{fig:example}
\end{figure}

\textbf{The comparison move types.} The board is held fixed and only the move changes to control for as many potential positional confounding variables as possible. Quiet moves are the model's
highest-logit moves that create no new attack on anything, including discovered attacks, give no check, capture nothing, and do not hang the piece. The best alternative is the model's highest
probability non-fork move. Blunders are the model's highest probability piece-hanging move. Each is picked once, at Elo 1500, and then swept across all nineteen Elos, so that the item being measured
cannot change from one to the next.

\begin{center}\rule{0.5\linewidth}{0.5pt}\end{center}

\section{Results}\label{results}

\subsection{A four for four monotonic depth migration for forks}\label{a-four-for-four-monotonic-depth-migration-for-forks}

We start by investigating how the skill conditioning aspect of the model affects fork circuitry following ablation experiments with them in other work (Litman, 2026c).

We first perform a head ablation sweep on 500 mined forks for each piece type, across Elo points spanning 700--2500 by intervals of 100. This occupies the majority of the range that Maia-3 was trained
on. We record how many of the Elo steps (e.g., 1100 to 1200) the center of mass increases on.

The table below reports on the subset of the 500 positions where the fork is the model's top move at every Elo, serving as a control that the move being measured is the move the model plays
throughout. A permutation test is run against the null hypothesis that Elo carries no information about the center of mass, shuffling the Elo labels within each position 1,000 times. The observed
shifts exceed the null (p \textless{} 0.001 for every piece).

{\def\LTcaptype{none} % do not increment counter
\begingroup\small\setlength{\tabcolsep}{4pt}
\begin{longtable}[]{@{}lllll@{}}
\toprule\noalign{}
piece & n & total ∆COM 700→2500 (layers) & Elo steps rising & permutation \emph{p} \\
\midrule\noalign{}
\endhead
\bottomrule\noalign{}
\endlastfoot
pawn & 261 & +0.519 ± 0.040 & 18/18 & \textless{} 0.001 \\
bishop & 243 & +0.547 ± 0.040 & 18/18 & \textless{} 0.001 \\
knight & 368 & +0.880 ± 0.028 & 18/18 & \textless{} 0.001 \\
rook & 238 & +0.522 ± 0.041 & 18/18 & \textless{} 0.001 \\
\end{longtable}
\endgroup
}

The full 100-step increments are in Section \ref{appendix}.

The results we obtain paint a consistent picture of a monotonic deepening of the heads' causal center of mass in all four pieces. Visualized:

\vspace{10pt}
\par\noindent\makebox[\linewidth][c]{\includegraphics[width=7.5in,keepaspectratio]{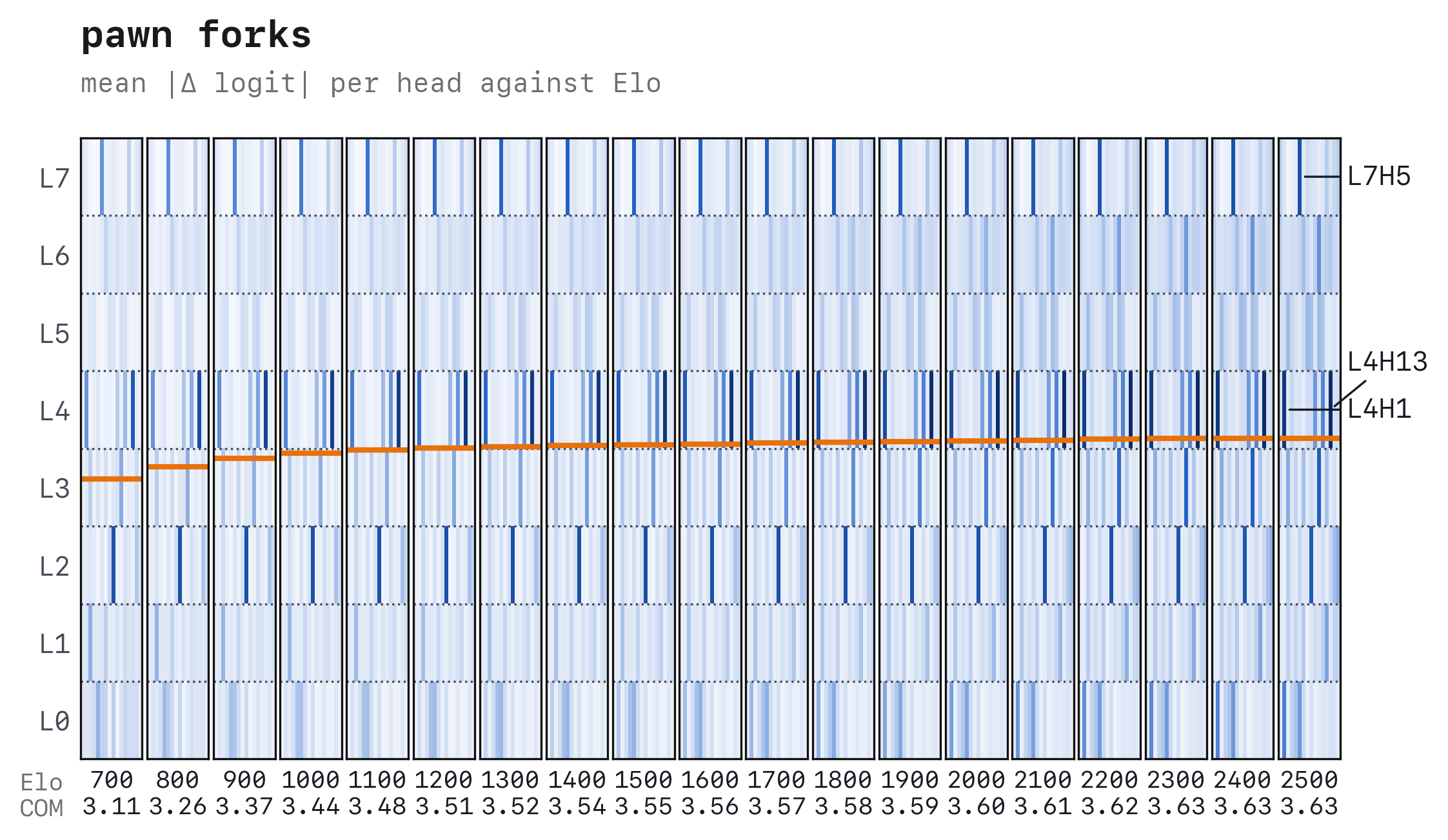}} \makebox[\linewidth][c]{\includegraphics[width=7.5in,keepaspectratio]{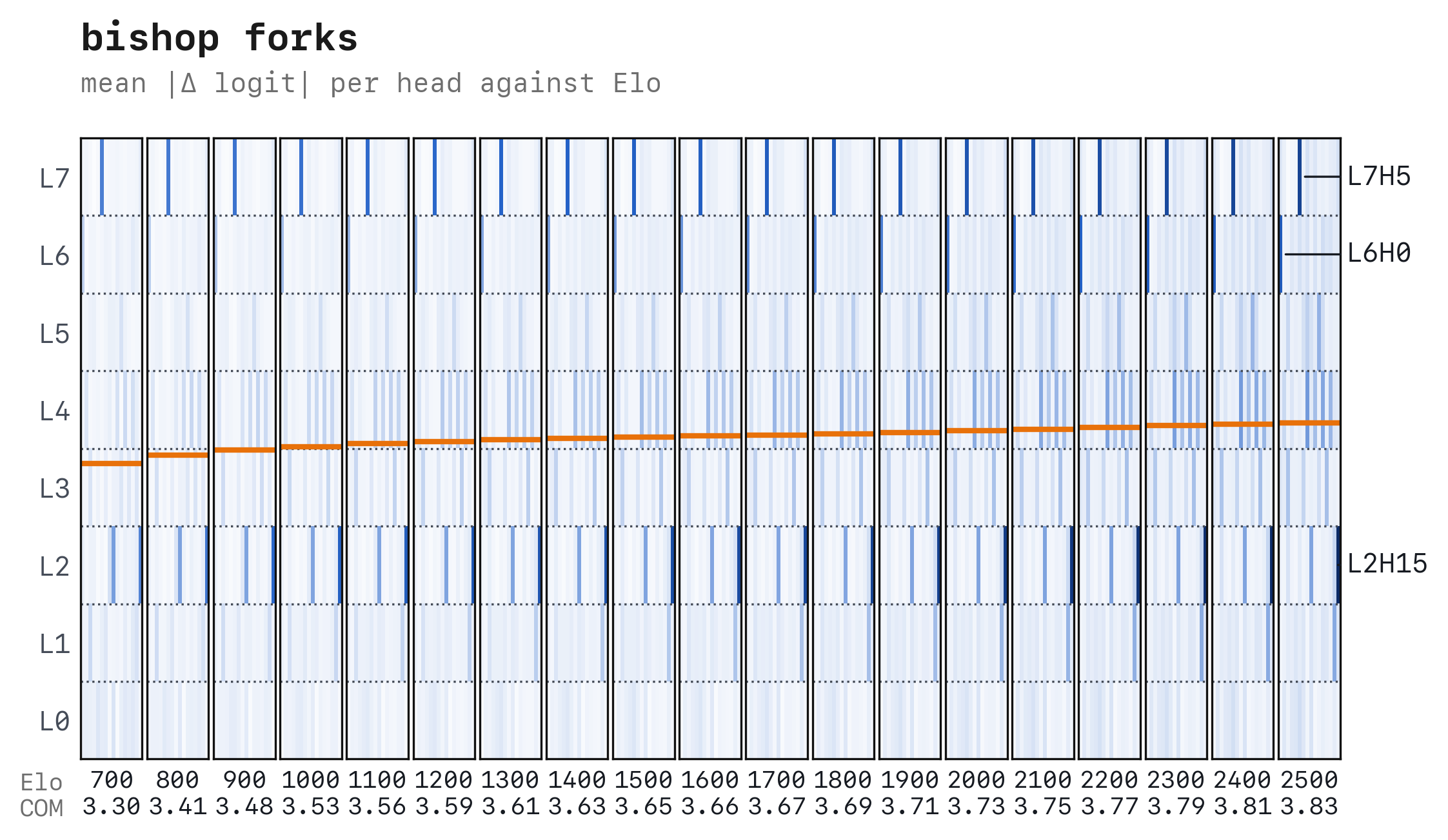}}
\par\noindent\makebox[\linewidth][c]{\includegraphics[width=7.5in,keepaspectratio]{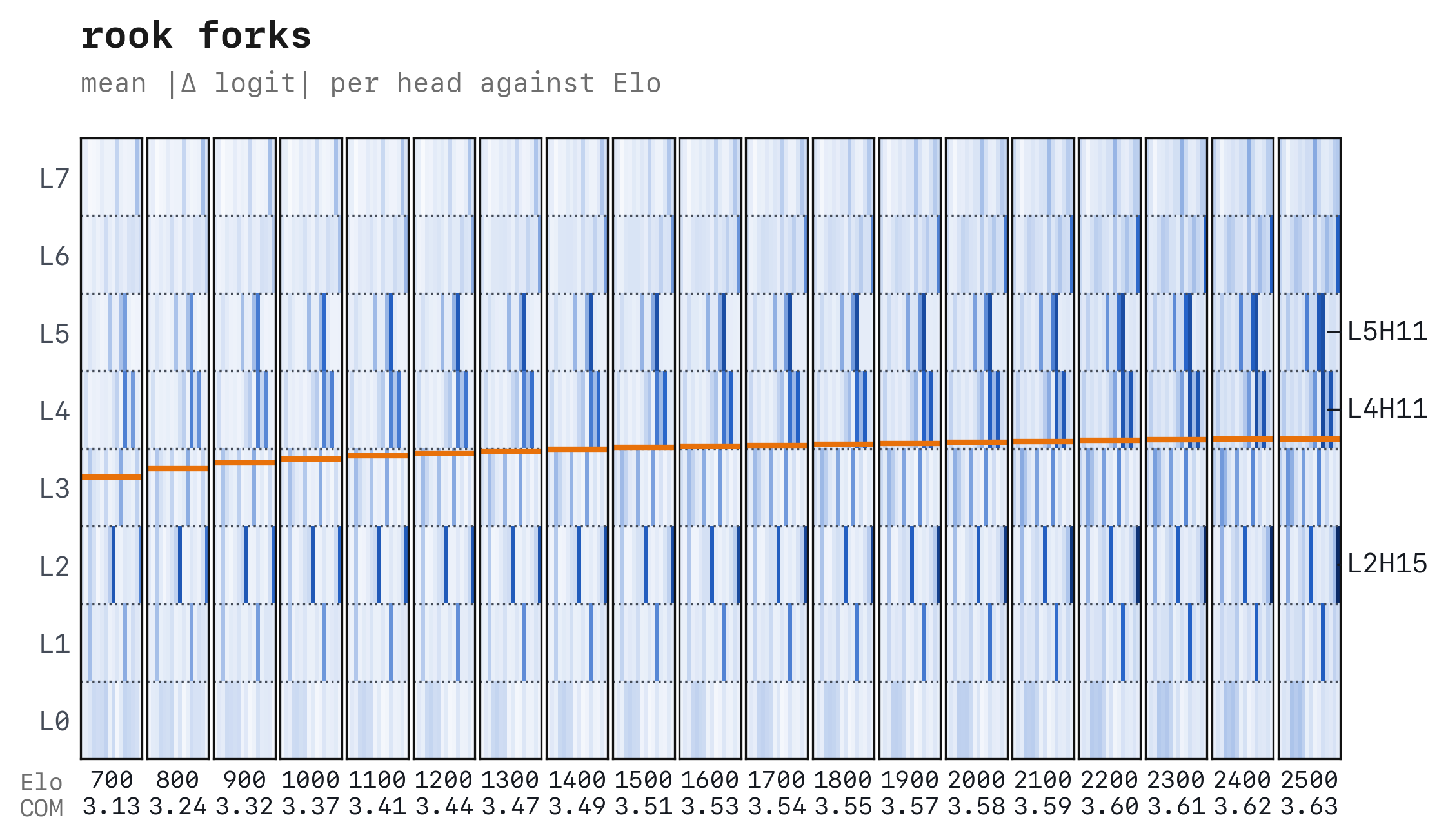}}
\begingroup\captionof{figure}{The remaining three pieces (the knight is Figure \ref{fig:ladder-knight}). Each panel shows the 16 heads per layer ``L'' with causal mass as brightness. So a brighter head means that ablating it
changes the move's logit more on average. This is plotted against Elo on the x axis. The orange line is the center of mass.
}\label{fig:ladders}\endgroup
\vspace{10pt}

We started with forks because of their unambiguousness and moderate complexity as a concept, but our results now raise the question: Does everything cause depth migration as we dial the skill up?

\subsection{Is the depth migration global?}\label{is-the-depth-migration-global}

To answer this question and search for structure within the migration, we ran the same head-ablation sweep on swathes of other data. We decided it was most fitting to first consider some
decompositions of a fork: the original ``fork'', the ``check-only'' without the queen attack, the ``queen attack only'' without the check, and another ``double-attack'' that hits two pieces except the
king or queen (Litman, 2026b). Since we cannot just remove a king from the board, we relocate the enemy king to the nearest safe square (trialed iteratively) such that it does not end up attacked and
the other target stays attacked. All edits must leave the board in a legal state and keep the mover uncapturable.

Below is the 2x2 surgery for the running example. The original move is highlighted and its favorability in Maia-3's policy is readable:

\begin{figure}
\centering
\makebox[\linewidth][c]{\includegraphics[width=\linewidth,keepaspectratio]{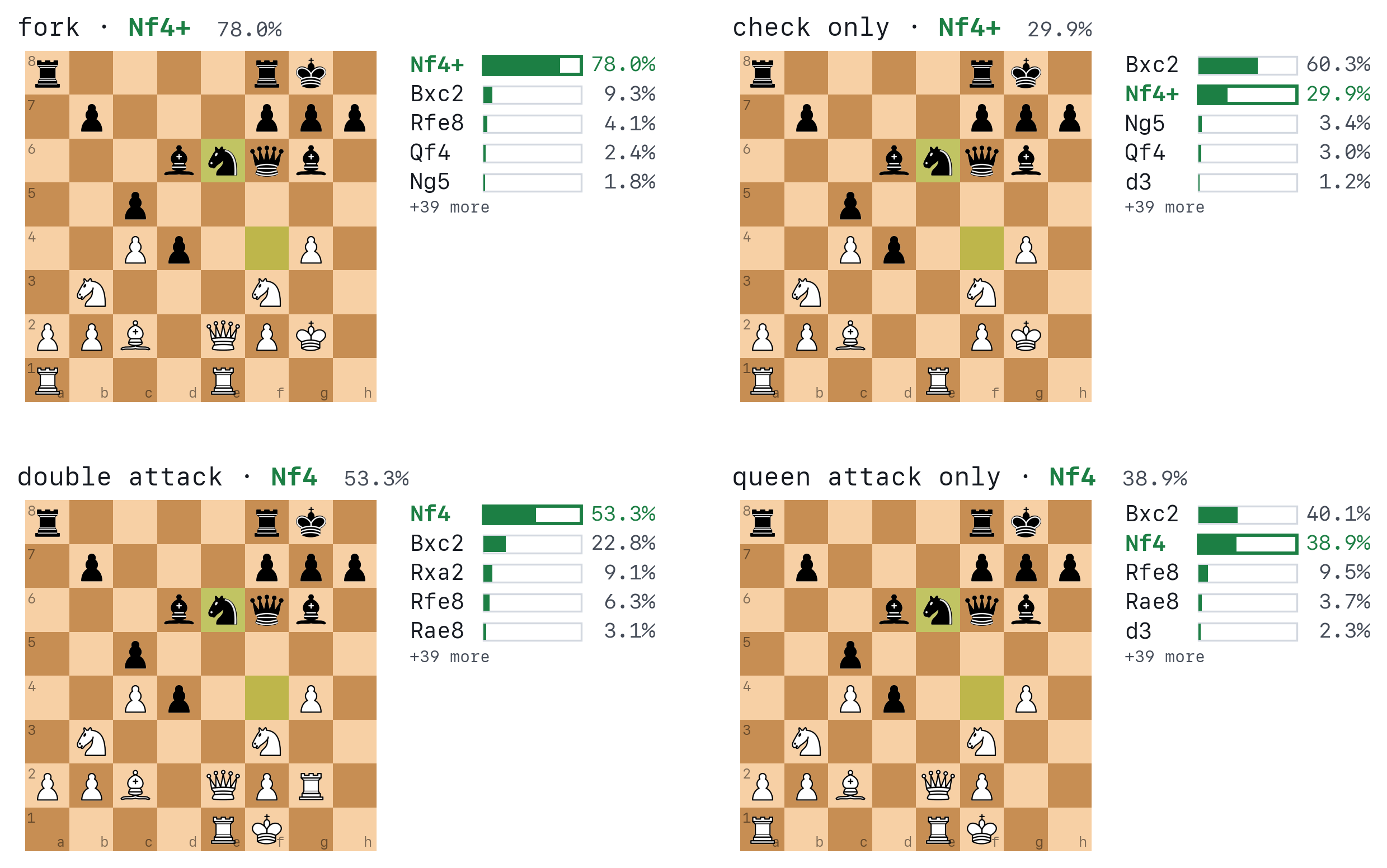}}
\caption{2x2 surgeries in the same position for ablation analysis. Fork and check-only above, double-attack and queen-attack-only below. The position stays relatively constant while only the content
of the move changes.}\label{fig:surgery}
\end{figure}

The fork decomposition data raises questions about composition and holisticity of chess features that we defer to later works.

We also ran head-ablation sweeps for the model's best alternative move and for its most preferred quiet move, to give a lower bound for the global movement.

\begin{figure}
\centering
\makebox[\linewidth][c]{\includegraphics[width=\linewidth,keepaspectratio]{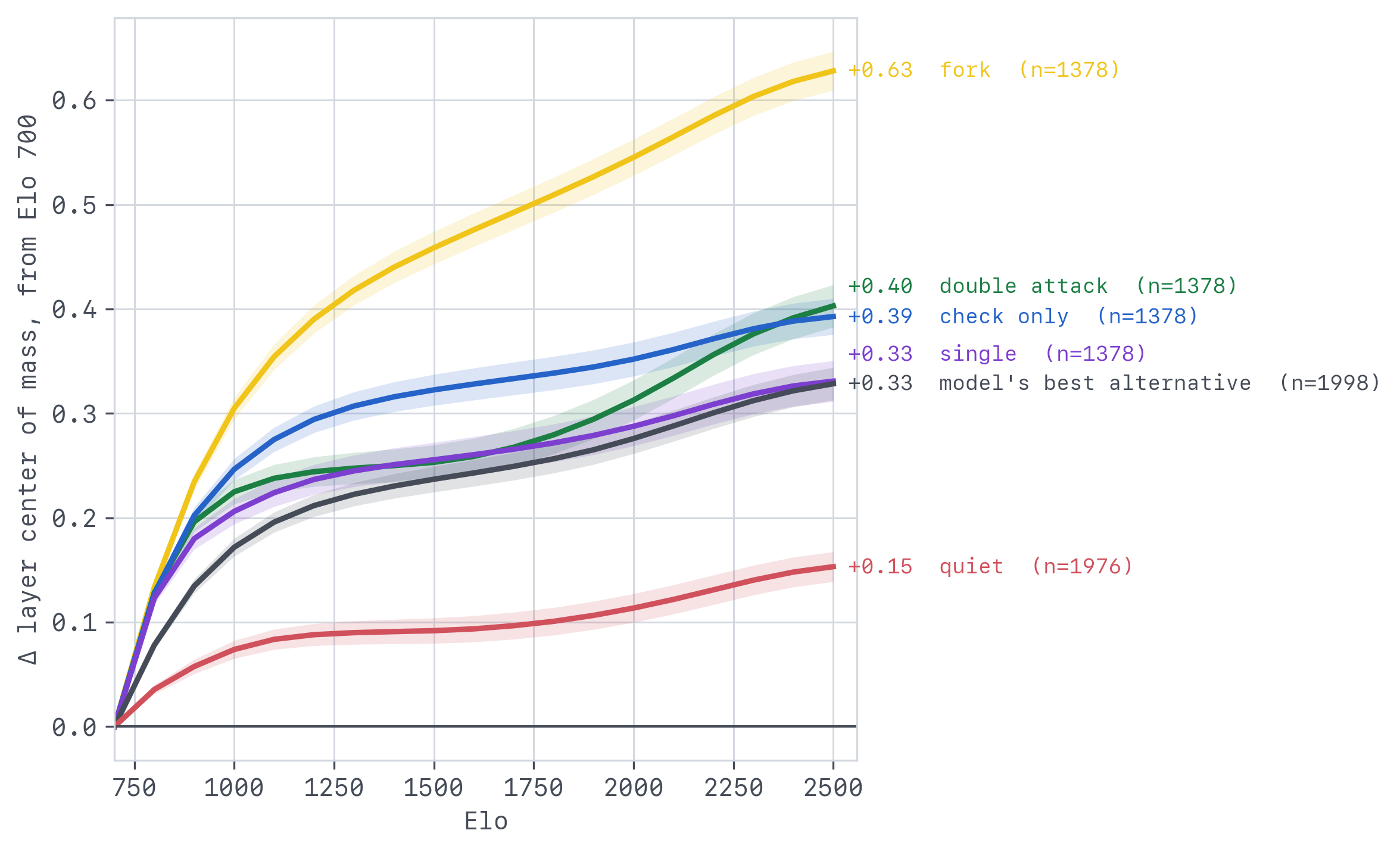}}
\caption{The causal mass for every move type we measured migrates deeper with Elo.}\label{fig:all-arms}
\end{figure}

Even the innocuous and inert quiet moves rely on later and later layers so the global effect is not primarily about tactics. Even outright blunders migrate, which we will discuss further in Section
\ref{controls}.

\subsection{Is there meaningful structure to the depth migration?}\label{is-there-meaningful-structure-to-the-depth-migration}

We looked for interpretable structure in the depth migration by piece and move. Drawing on work by Jenner et al.~(2024), we sought to examine depth migration in a variety of pre-labeled chess puzzles,
hypothesizing that perhaps the number of steps needed to look ahead, a temporary material loss, or some other tactic might reveal a new interesting pattern of causal center of mass migration.

Using a Lichess.org supplied puzzle database, we ran a simplified head-ablation sweep with a coarser Elo partition on a list of puzzle concepts like ``double attack'' and ``checkmate in 3.''

The number reported for each concept is the excess migration of the puzzle's solution move compared to the migration of the model's best other move in that same position. Our knight fork's excess is
+0.513. Its raw +0.850 migration is that much more than the best other move's +0.337 layers.

{\def\LTcaptype{none} % do not increment counter
\begingroup\small\setlength{\tabcolsep}{4pt}
\begin{longtable}[]{@{}lllll@{}}
\toprule\noalign{}
concept & n & excess & 95\% CI & description \\
\midrule\noalign{}
\endhead
\bottomrule\noalign{}
\endlastfoot
smotheredMate & 350 & +0.300 & {[}+0.258, +0.344{]} & knight-delivered checkmate \\
discoveredAttack & 350 & +0.279 & {[}+0.220, +0.344{]} & uncovers an attack by moving a piece \\
hangingPiece & 350 & +0.235 & {[}+0.182, +0.286{]} & \\
fork & 350 & +0.202 & {[}+0.156, +0.249{]} & any piece that attacks any two targets \\
discoveredCheck & 350 & +0.178 & {[}+0.133, +0.226{]} & \\
mateIn1 & 350 & +0.154 & {[}+0.112, +0.196{]} & guaranteed checkmate in one move \\
mateIn3 & 350 & +0.115 & {[}+0.062, +0.164{]} & \\
doubleCheck & 350 & +0.111 & {[}+0.070, +0.159{]} & \\
mateIn2 & 350 & +0.098 & {[}+0.053, +0.143{]} & \\
veryLong & 350 & +0.097 & {[}+0.044, +0.145{]} & a long solution but with no checkmate \\
skewer & 350 & +0.087 & {[}+0.038, +0.135{]} & \\
pin & 350 & +0.084 & {[}+0.039, +0.131{]} & \\
sacrifice & 350 & +0.012 & {[}−0.037, +0.063{]} & not distinguishable from zero \\
\textbf{our royal knight fork} & 500 & \textbf{+0.513} & {[}+0.480, +0.546{]} & \\
\end{longtable}
\endgroup
}

None of the puzzle concepts migrated more than our king queen fork, and puzzle difficulty did not correlate with depth migration considering mateIn1, mateIn2 and mateIn3 have overlapping confidence
intervals. Also, veryLong proved indistinguishable from them as well. The five deepest migrations include the smothered mate (a knight checkmate) and three concepts involving two pieces: discovered
attack, discovered check, and the general fork. Those are each related in a way to our knight fork. Sacrifice is the only concept whose interval covers zero. So puzzle difficulty does not predict how
far a move migrates, but the make-up of the move does.

The next table suggests that the model organizes its ontology by piece type rather than move type. Consider the depth migration of each decomposition of a fork across all pieces. All four columns use
the surgery of Section \ref{is-the-depth-migration-global}, which changes only the class being tested.

{\def\LTcaptype{none} % do not increment counter
\begingroup\small\setlength{\tabcolsep}{4pt}
\begin{longtable}[]{@{}llllll@{}}
\toprule\noalign{}
piece & n & fork & check only & double attack & queen attack only \\
\midrule\noalign{}
\endhead
\bottomrule\noalign{}
\endlastfoot
pawn & 355 & +0.548 ± 0.037 & +0.537 ± 0.033 & +0.482 ± 0.042 & +0.510 ± 0.041 \\
bishop & 385 & +0.533 ± 0.034 & +0.284 ± 0.032 & +0.256 ± 0.035 & +0.183 ± 0.034 \\
knight & 392 & +0.864 ± 0.026 & +0.400 ± 0.033 & +0.558 ± 0.037 & +0.342 ± 0.036 \\
rook & 246 & +0.516 ± 0.040 & +0.343 ± 0.033 & +0.273 ± 0.037 & +0.286 ± 0.036 \\
\end{longtable}
\endgroup
}

The pieces differ not only in magnitude. The pawn's depth is hardly fork-driven since forks and check only values are indistinguishable. The knight fork's shift (+0.86) exceeds the check-only and
queen-attack-only shifts combined (+0.40 + 0.34), which is consistent with the fork feature being more than the sum of its parts. But as mentioned in previous work, transcoders would shed considerable
light on this (Litman, 2026b).

\subsection{Controls}\label{controls}

Two alternative explanations need ruling out. We frame them as objections.

\textbf{Objection:} ``What is being measured is not related to the moves themselves, such as forks; it is a function of the model's confidence in a strong move, which by design increases with Elo.''

If depth were a function of confidence, a move whose probability decreases as Elo rises should migrate shallower, or possibly not at all. This does not occur.

The best alternative move in the positions where the model plays the fork loses probability as Elo rises (from 0.13--0.15 at Elo 700 to 0.06--0.09 at 2500, depending on the piece) and still deepens by
+0.22 ± 0.04 (pawn), +0.23 ± 0.04 (bishop), +0.29 ± 0.04 (knight) and +0.26 ± 0.05 (rook) layers. Outright blunders, the model's most favored piece-hanging move in the same positions, have the same
trend with a much larger effect. For pawns, knights, and rooks they more than halve in probability and for bishops they decrease by a third. Yet each blunder deepens by +0.209 ± 0.031 (pawn), +0.249 ±
0.030 (bishop), +0.318 ± 0.030 (knight) and +0.218 ± 0.029 (rook) layers. Both the best alternative moves and the blunders were mined from the n=500 per piece set.

This shows that the global depth migration effect still occurs as the model likes a move less and less. It is not a function of confidence.

\textbf{Objection:} ``The ablation metric itself is a lucky choice and the depth migration does not generalize.''

Ablation of attention heads is a foundational method for determining where computation happens in a transformer. However, our results are corroborated by another metric:

We measured with logit-lens where the logit for a fork surpassed 75\% of its final logit score of the forward pass. We used a logit threshold not a rank threshold as inspired by Zhang and Nanda (2023)
who note that a rank metric is confounded by how other outputs are suppressed and so is not as reliable at localizing a specific computation. Our high fraction was chosen because early in the model
there is a lot of uncertainty while the network is still parsing the move.

\begin{figure}
\centering
\makebox[\linewidth][c]{\includegraphics[width=\linewidth,keepaspectratio]{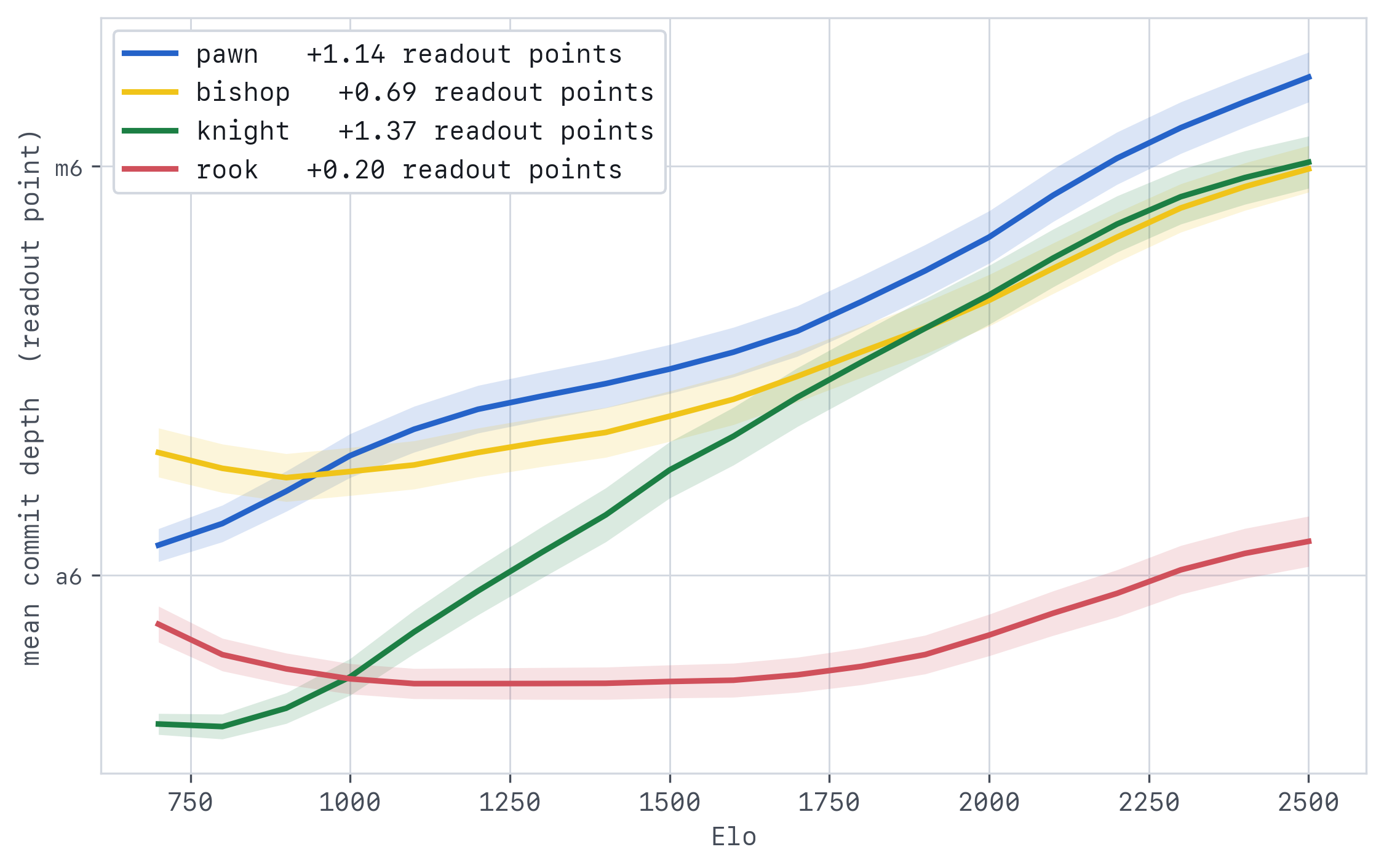}}
\caption{Ablation-free metric of depth migration: the move crosses its logit threshold later as skill is dialed up.}\label{fig:commit}
\end{figure}

Notice the near monotonic rise across all pieces as in the head-ablation sweeps.

\subsection{A first pass at a mechanistic explanation}\label{a-first-pass-at-a-mechanistic-explanation}

One might predict that the migration would be to shallower direction as skill increased. In a neural network, the more layers there are after a feature is computed, the more opportunities there are to
use that feature in subsequent computations. So a more advanced and skilled network should learn features like forks earlier on to reuse them in later layers (Princeton University Professor Tom
Griffiths, personal communication, September 2026).

We showed the opposite in our specific data. That direction matches the human result: accumulated expertise directly increases a player's planning depth, and the increase in planning depth accounts
for much of the gain in skill (van Opheusden et al., 2023).

Mechanistic interpretations are harder than describing mathematical relationships, but we can make a first pass at one by analyzing specific heads' trajectories as Elo is dialed up.

\begin{figure}
\centering
\makebox[\linewidth][c]{\includegraphics[width=\linewidth,keepaspectratio]{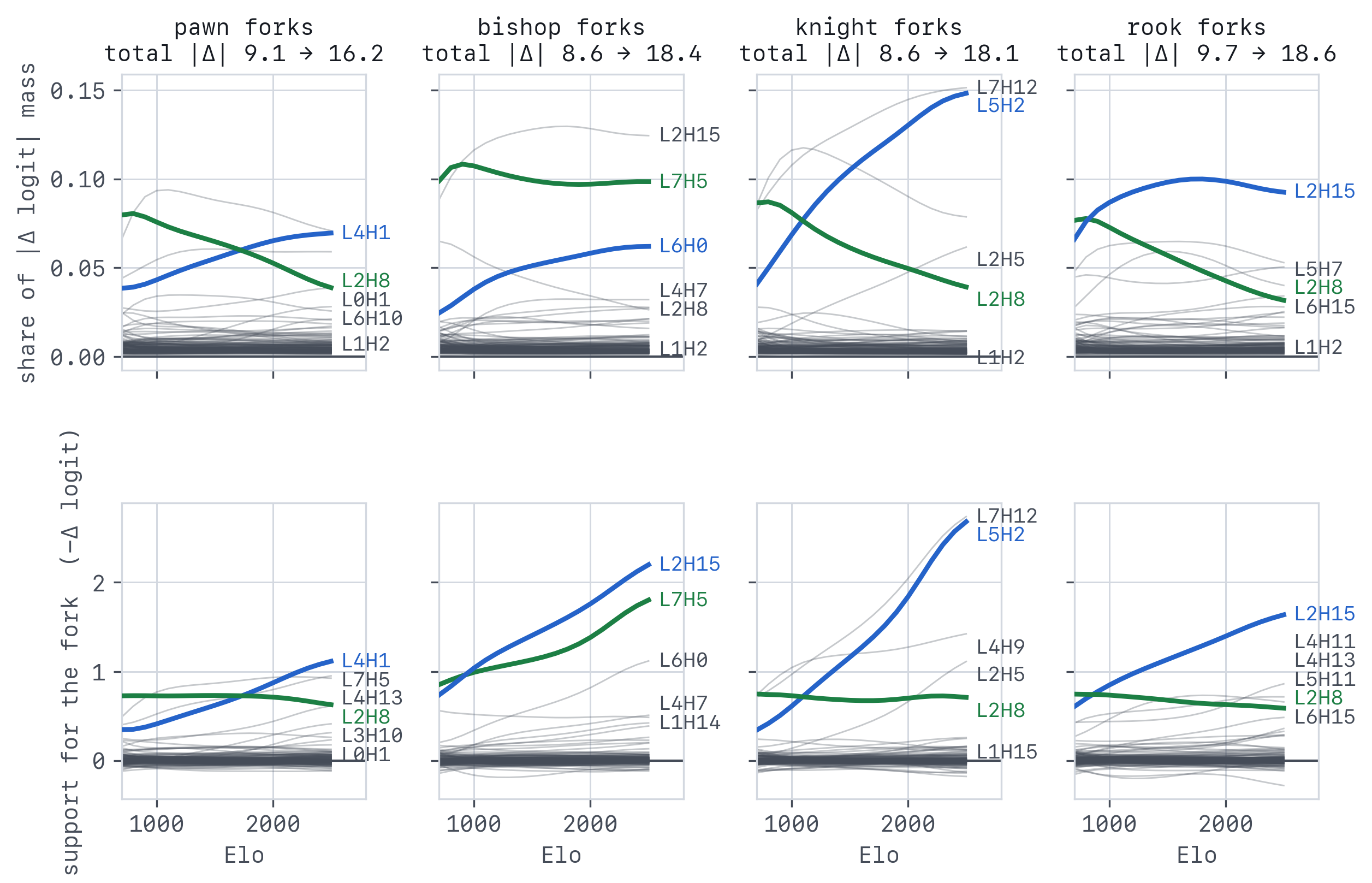}}
\caption{Fork head trajectories. Relative mass above, absolute mass below. Green denotes the head with the most causal mass at the first Elo point and blue denotes the head that increases in causal
mass the most. The other heads with the largest changes are labelled in grey.}\label{fig:trajectories}
\end{figure}

Consider, to start, L2H8. Its absolute contribution is more or less constant for each piece across Elos, and lies between −0.5 and −0.75 logit. Because the total causal mass increases with Elo, this
head starts as an important carrier of forks and is relied upon less and less in relative terms (note its decline in the relative-mass plots). It never loses its absolute signal, which matches the
linear nature of the Elo parameter.

Now consider L5H2 and L7H12 on the knight's plots. They contribute more and more as skill increases, growing four to eight times in absolute contribution and ending up responsible for 30\% of the
total causal mass on knight forks at Elo 2500.

A candidate interpretation of the above facts is that L2H8, being located on an early layer, is an early specialization head that all forking pieces rely on in low skill settings. As skill
conditioning increases, the network relies less and less on the computations from that early layer, though still a constant amount, and rapidly recruits more specialized heads for more precise
computations. The layer 2 head could be computing a reusable primitive and the later heads could be piece specific or deal with more abstract features of a position.

As mentioned earlier, attention maps from a square correspond to the 8x8 board so we can naturally view the attention heads on any position, with color representing the strength of the softmaxed
semantic plus GAB attention maps.

On a randomly generated position from Maia-3 self-play, we draw an attention atlas with chessformer\_lens for the heads above

\begin{figure}
\centering
\makebox[\linewidth][c]{\includegraphics[width=\linewidth,keepaspectratio]{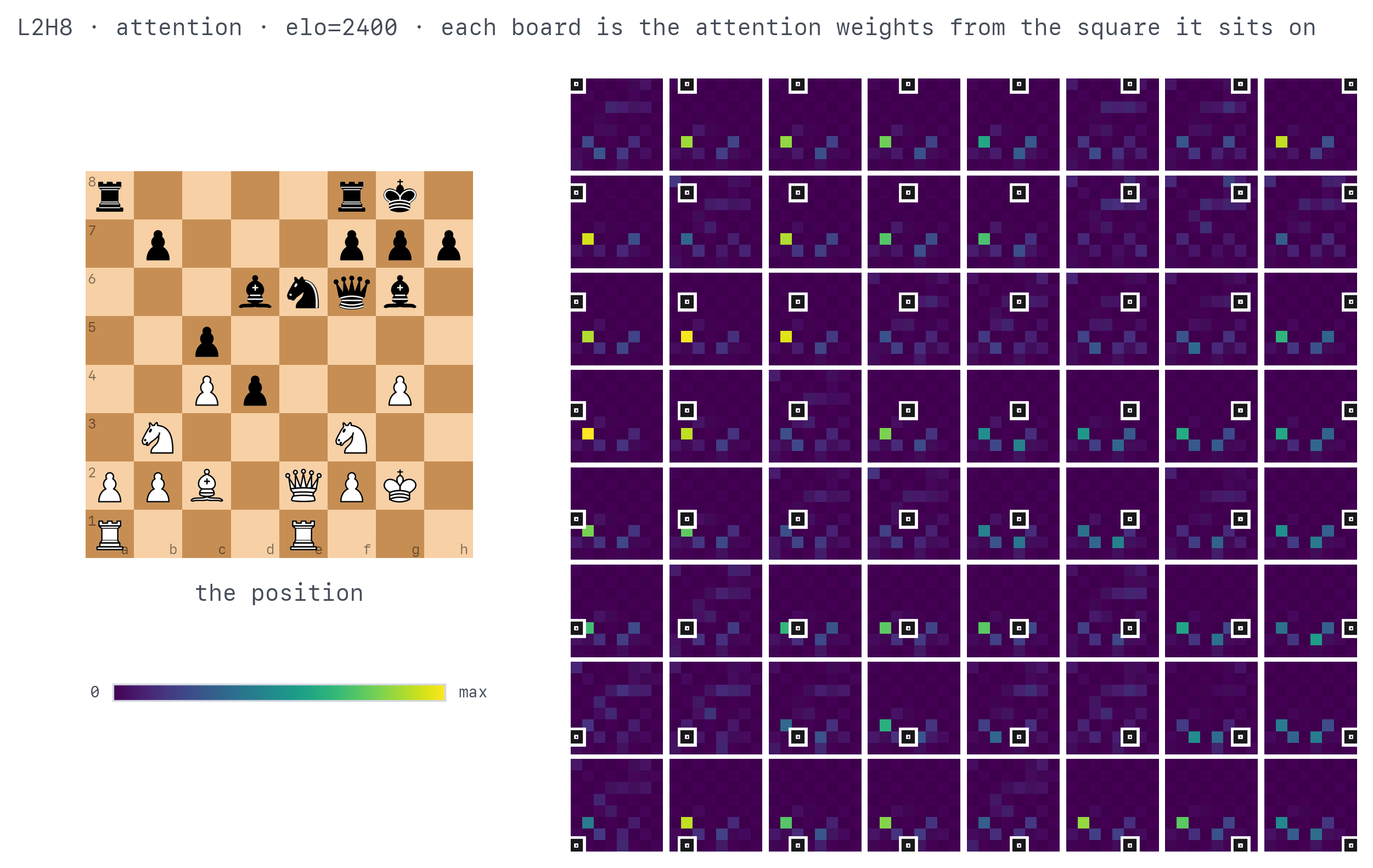}}
\caption{L2H8 seems to attend from a square to an assortment of high value pieces that are diagonally near each other. It varies little by position.}\label{fig:atlas-l2h8}
\end{figure}

\begin{figure}
\centering
\makebox[\linewidth][c]{\includegraphics[width=\linewidth,keepaspectratio]{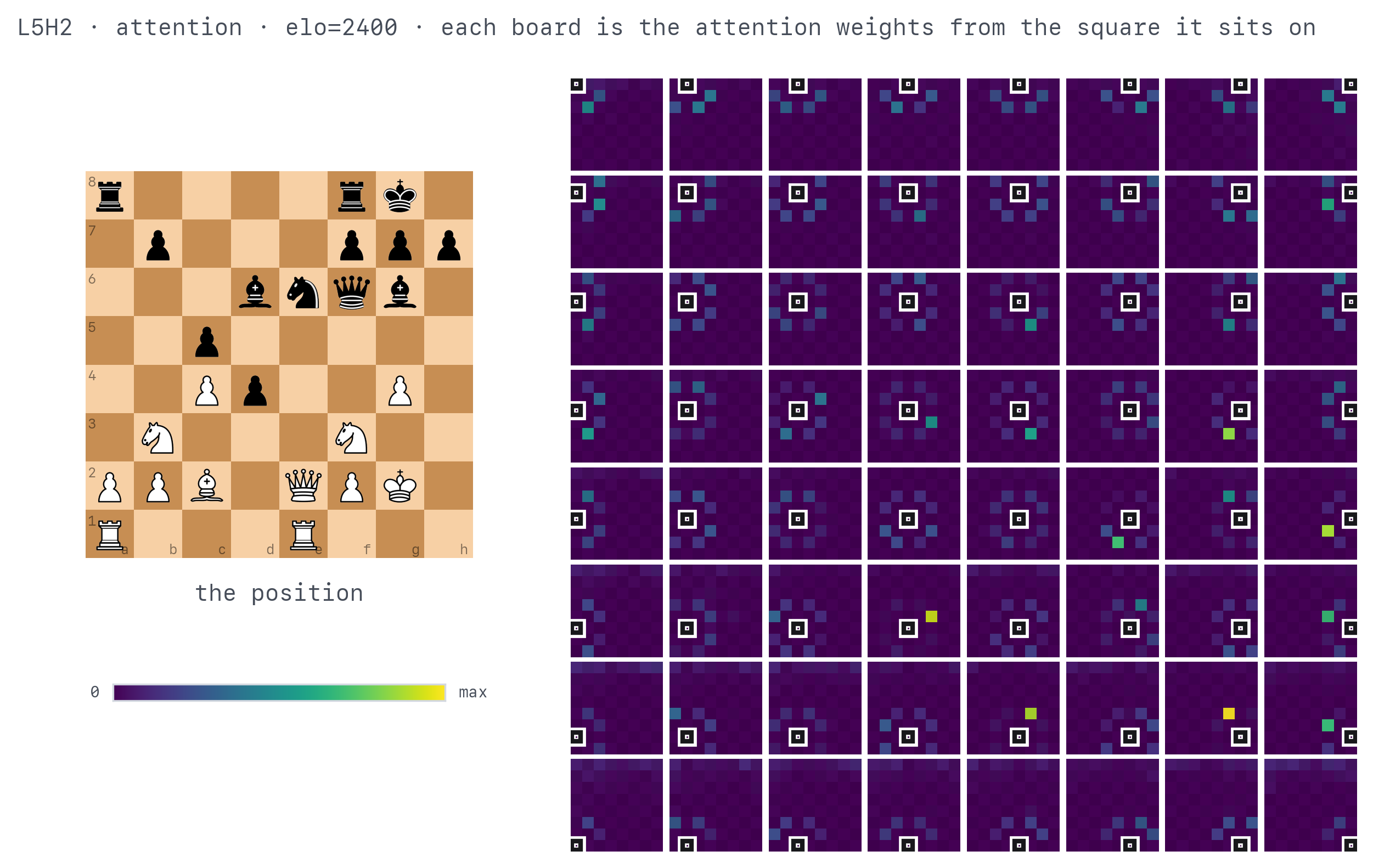}}
\caption{L5H2 seems to attend precisely in the pattern of knight geometry, and fire most from a high value piece's square to a square from which a knight move would fork it and
another.}\label{fig:atlas-l5h2}
\end{figure}

\begin{figure}
\centering
\makebox[\linewidth][c]{\includegraphics[width=\linewidth,keepaspectratio]{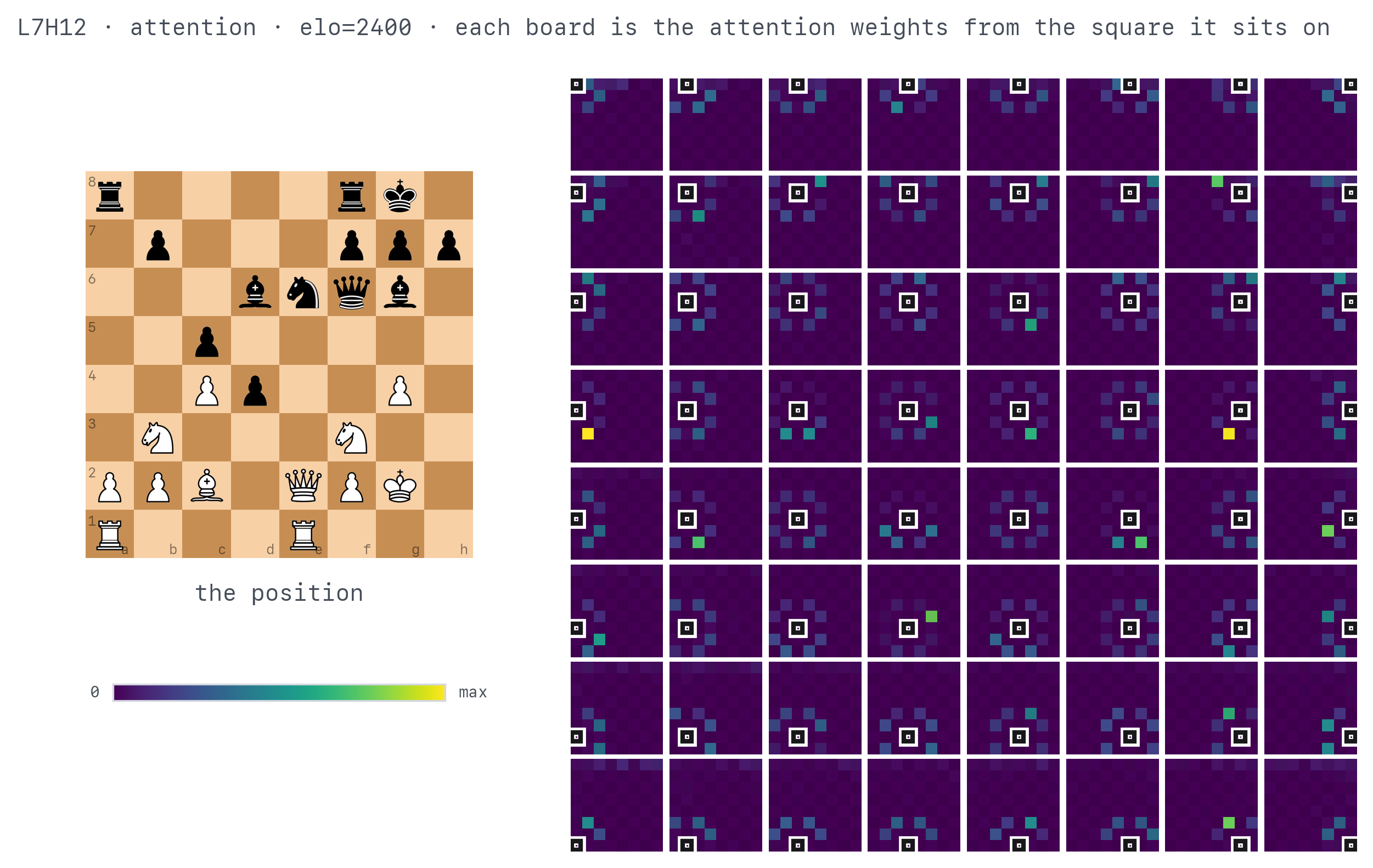}}
\caption{L7H12 does something similar to L5H2 but more diffusely, with firings that are more difficult to explain and could reflect a more complicated computation two layers
deeper.}\label{fig:atlas-l7h12}
\end{figure}

Considering the analysis of those three heads, the skill conditioning that gives rise to depth migration is (at least for knights) likely not a result of computations being routed to wholly new
circuits. It is a result of the model relying less on shallower heads' primitive calculations like ``where are the forkable pieces'' and more on deeper heads with more specialized, geometry-specific
computations.

\begin{center}\rule{0.5\linewidth}{0.5pt}\end{center}

\section{Limitations}\label{limitations}

\begin{itemize}
\tightlist
\item
  We use one Maia model size at 23 million parameters (though initial results suggest the finding holds in the 5M and 79M models).
\item
  The mining predicates are improvable. They were iteratively written, for instance by realizing that we might be eliding discovered checks and adding a conditional for them, so further improvements
  may yield cleaner results. This is an especially important area to focus on since the predicates form the basis of our sampling and thus are upstream of every figure and datum.
\item
  The logit lens reads everything in the policy head's basis. This means that a feature represented outside the directions read by the head's from and to-square projections will not be detected,
  constraining our commit-depth results to a subspace of decodable representations only (nostalgebraist, 2020).
\end{itemize}

\textbf{Future questions}

\begin{itemize}
\tightlist
\item
  What do the recruited deeper heads compute, and how do suppression heads figure into the model's circuitry? We take this up in forthcoming work.
\item
  Does this result hold in other models with an analogue to skill conditioning, inside and outside of chess?
\item
  Can transcoders or SAEs determine whether the fork feature is more than the sum of its parts?
\end{itemize}

\section{Acknowledgements}\label{acknowledgements}

I am grateful to Francis Crick Chair Professor Terry Sejnowski for feedback on this write-up and for his experienced intuitions about the results and how to best square them with larger scientific
ideas. I thank Professor Tom Griffiths for generous discussions of these ideas at Princeton, and for his introductions to colleagues who have since offered to collaborate.

This work was carried out independently without external funding. The author conceived the question, built the chessformer\_lens library, ran the experiments, and wrote the paper; ``we'' is used
editorially.

\section{Appendix}\label{appendix}

\begin{itemize}
\tightlist
\item
  \textbf{Notebook to recreate the results:} \url{https://github.com/David-31415/maia-depth-migration}
\item
  \textbf{Library used:} \url{https://github.com/chessformer-lens/chessformer_lens}
\end{itemize}

\textbf{Table A1.} Change in the causal center of mass at each Elo step, on the positions where the fork is the model's top move at every Elo (95\% CI over positions).

{\def\LTcaptype{none} % do not increment counter
\begingroup\small\setlength{\tabcolsep}{4pt}
\begin{longtable}[]{@{}lllll@{}}
\toprule\noalign{}
Elo & pawn ∆COM & bishop ∆COM & knight ∆COM & rook ∆COM \\
\midrule\noalign{}
\endhead
\bottomrule\noalign{}
\endlastfoot
700→800 & +0.146 ± 0.010 & +0.108 ± 0.010 & +0.169 ± 0.007 & +0.111 ± 0.009 \\
800→900 & +0.107 ± 0.009 & +0.069 ± 0.009 & +0.145 ± 0.006 & +0.082 ± 0.008 \\
900→1000 & +0.067 ± 0.007 & +0.046 ± 0.008 & +0.107 ± 0.005 & +0.056 ± 0.007 \\
1000→1100 & +0.040 ± 0.005 & +0.035 ± 0.006 & +0.077 ± 0.004 & +0.041 ± 0.006 \\
1100→1200 & +0.024 ± 0.004 & +0.029 ± 0.005 & +0.057 ± 0.003 & +0.034 ± 0.005 \\
1200→1300 & +0.017 ± 0.003 & +0.023 ± 0.004 & +0.044 ± 0.003 & +0.030 ± 0.004 \\
1300→1400 & +0.012 ± 0.003 & +0.019 ± 0.003 & +0.035 ± 0.002 & +0.024 ± 0.003 \\
1400→1500 & +0.010 ± 0.003 & +0.016 ± 0.003 & +0.030 ± 0.002 & +0.019 ± 0.003 \\
1500→1600 & +0.010 ± 0.003 & +0.015 ± 0.003 & +0.027 ± 0.002 & +0.016 ± 0.003 \\
1600→1700 & +0.010 ± 0.004 & +0.015 ± 0.003 & +0.025 ± 0.002 & +0.013 ± 0.003 \\
1700→1800 & +0.010 ± 0.004 & +0.017 ± 0.004 & +0.023 ± 0.003 & +0.012 ± 0.003 \\
1800→1900 & +0.011 ± 0.004 & +0.019 ± 0.004 & +0.022 ± 0.003 & +0.012 ± 0.003 \\
1900→2000 & +0.012 ± 0.004 & +0.022 ± 0.004 & +0.022 ± 0.003 & +0.012 ± 0.004 \\
2000→2100 & +0.013 ± 0.004 & +0.023 ± 0.004 & +0.022 ± 0.003 & +0.013 ± 0.004 \\
2100→2200 & +0.013 ± 0.004 & +0.026 ± 0.004 & +0.022 ± 0.003 & +0.014 ± 0.004 \\
2200→2300 & +0.010 ± 0.003 & +0.026 ± 0.004 & +0.021 ± 0.003 & +0.013 ± 0.004 \\
2300→2400 & +0.005 ± 0.003 & +0.022 ± 0.003 & +0.017 ± 0.002 & +0.012 ± 0.003 \\
2400→2500 & +0.002 ± 0.003 & +0.017 ± 0.003 & +0.013 ± 0.002 & +0.008 ± 0.003 \\
\textbf{700→2500} & \textbf{+0.519 ± 0.040} & \textbf{+0.547 ± 0.040} & \textbf{+0.880 ± 0.028} & \textbf{+0.522 ± 0.041} \\
Elo steps rising & 18/18, \emph{p} \textless{} 0.001 & 18/18, \emph{p} \textless{} 0.001 & 18/18, \emph{p} \textless{} 0.001 & 18/18, \emph{p} \textless{} 0.001 \\
n & 261 & 243 & 368 & 238 \\
\end{longtable}
\endgroup
}

\section{References}\label{references}
\setlength{\parskip}{10pt}\setlength{\parindent}{0pt}\everypar{\hangindent=1.5em\relax}

Hu, Y., Zhou, C., and Zhang, M. (2025). What Affects the Effective Depth of Large Language Models? arXiv:2512.14064.

Jenner, E., Kapur, S., Georgiev, V., Allen, C., Emmons, S., and Russell, S. (2024). Evidence of Learned Look-Ahead in a Chess-Playing Neural Network. Advances in Neural Information Processing Systems
37. arXiv:2406.00877.

Karvonen, A. (2024). Emergent World Models and Latent Variable Estimation in Chess-Playing Language Models. Conference on Language Modeling (COLM). arXiv:2403.15498.

Litman, D. (2026b). Fork Around and Find Out Part 2: One Head does the Summing. LessWrong, 15 July 2026.
\url{https://www.lesswrong.com/posts/6reCnPYeopThEFQxN/fork-around-and-find-out-part-2-one-head-does-the-summing}

Litman, D. (2026c). Fork Around and Find Out Part 3: Interpreting the Knight Auditor. LessWrong, 17 August 2026.
\url{https://www.lesswrong.com/posts/Cke4aTXGB7aG8zCsx/fork-around-and-find-out-part-3-interpreting-the-knight}

Litman, D. (2026d). chessformer\_lens: an interpretability lens for square-token chess transformers. Version 0.3.0, 27 August 2026. \url{https://doi.org/10.5281/zenodo.21877655}

McGrath, T., Kapishnikov, A., Tomašev, N., Pearce, A., Wattenberg, M., Hassabis, D., Kim, B., Paquet, U., and Kramnik, V. (2022). Acquisition of Chess Knowledge in AlphaZero. Proceedings of the
National Academy of Sciences 119(47). arXiv:2111.09259.

McIlroy-Young, R., Sen, S., Kleinberg, J., and Anderson, A. (2020). Aligning Superhuman AI with Human Behavior: Chess as a Model System. Proceedings of KDD 2020. arXiv:2006.01855.

Monroe, D., Eilender, D., Chalmers, A., Tang, Z., and Anderson, A. (2026). Chessformer: A Unified Architecture for Chess Modeling. International Conference on Learning Representations (ICLR).
arXiv:2605.19091.

nostalgebraist (2020). interpreting GPT: the logit lens. LessWrong.

Olah, C., Cammarata, N., Schubert, L., Goh, G., Petrov, M., and Carter, S. (2020). Zoom In: An Introduction to Circuits. Distill 5(3). \url{https://distill.pub/2020/circuits/zoom-in/}

Ruoss, A., Delétang, G., Medapati, S., Grau-Moya, J., Wenliang, L. K., Catt, E., Reid, J., and Genewein, T. (2024). Grandmaster-Level Chess Without Search. Advances in Neural Information Processing
Systems 37, Datasets and Benchmarks Track. arXiv:2402.04494.

Shannon, C. E. (1950). Programming a Computer for Playing Chess. The London, Edinburgh, and Dublin Philosophical Magazine and Journal of Science 41(314), 256--275.

Silver, D., Hubert, T., Schrittwieser, J., Antonoglou, I., Lai, M., Guez, A., Lanctot, M., Sifre, L., Kumaran, D., Graepel, T., Lillicrap, T., Simonyan, K., and Hassabis, D. (2017). Mastering Chess
and Shogi by Self-Play with a General Reinforcement Learning Algorithm. arXiv:1712.01815.

Tang, Z., Jiao, D., McIlroy-Young, R., Kleinberg, J., Sen, S., and Anderson, A. (2024). Maia-2: A Unified Model for Human-AI Alignment in Chess. Advances in Neural Information Processing Systems 37.
arXiv:2409.20553.

van Opheusden, B., Kuperwajs, I., Galbiati, G., Bnaya, Z., Li, Y., and Ma, W. J. (2023). Expertise increases planning depth in human gameplay. Nature 618, 1000--1005.

Vaswani, A., Shazeer, N., Parmar, N., Uszkoreit, J., Jones, L., Gomez, A. N., Kaiser, Ł., and Polosukhin, I. (2017). Attention Is All You Need. Advances in Neural Information Processing Systems 30.
arXiv:1706.03762.

Zhang, F., and Nanda, N. (2023). Towards Best Practices of Activation Patching in Language Models: Metrics and Methods. arXiv:2309.16042.

\end{document}